\documentclass{article}
\usepackage[utf8]{inputenc}
\usepackage{fullpage}
\usepackage{amsmath}
\usepackage{amsfonts}
\usepackage{amssymb}
\usepackage{graphicx}  % DO NOT CHANGE THIS
\usepackage{caption}  % DO NOT CHANGE THIS
\usepackage{subcaption}
\usepackage{float}

\title{Active Predictive Coding Networks}
\author{gklezd }
\date{December 2021}

\begin{document}

\title{
Active Predictive Coding Networks: A Neural Solution to the Problem of Learning Reference Frames and Part-Whole Hierarchies}
\author{Dimitrios C. Gklezakos and Rajesh P. N. Rao\\
\normalsize Paul G. Allen School of Computer Science and Engineering\\
\normalsize University of Washington, Seattle, USA\\
\texttt{\normalsize gklezd@cs.washington.edu}, \texttt{\normalsize rao@cs.washington.edu}
}
\date{}

\maketitle
\begin{abstract}We introduce Active Predictive Coding Networks (APCNs), a new class of neural  networks that solve a major problem posed by Hinton and others in the fields of artificial intelligence and brain modeling: how can neural networks learn intrinsic reference frames for objects and parse visual scenes into part-whole hierarchies by dynamically allocating nodes in a parse tree? APCNs address this problem by using a novel combination of ideas: (1) hypernetworks are used for dynamically generating recurrent neural networks that predict parts and their locations within intrinsic reference frames conditioned on higher object-level embedding vectors, and (2) reinforcement learning is used in conjunction with backpropagation for end-to-end learning of model parameters. The APCN architecture lends itself naturally to multi-level hierarchical learning and is closely related to predictive coding models of cortical function. Using the MNIST, Fashion-MNIST and Omniglot datasets, we demonstrate that APCNs can (a) learn to parse images into part-whole hierarchies, (b) learn compositional representations, and (c) transfer their knowledge to unseen classes of objects. With their ability to dynamically generate parse trees with part locations for objects, APCNs offer a new framework for explainable AI that leverages advances in deep learning while retaining interpretability and compositionality.
\end{abstract}

\section{Introduction}
% Motivation and related work  
Deep convolutional neural networks have enabled path-breaking advances in visual classification problems in recent years \cite{NIPS2012_c399862d} but they suffer from a fundamental shortcoming: they do not preserve positional information about extracted features. Even though they may correctly classify an image, they are unable to explain the images they classify in the way humans do: in terms of objects, their locations in a scene, the parts of an object and the locations of these parts within the object, etc. This lack of interpretability of deep neural networks has prompted a search for alternate models that are inspired by how humans represent objects in terms of part-whole hierarchies and use compositionality of parts to explain new objects. For example, Hinton and colleagues \cite{sabour2017dynamic,NEURIPS2019_2e0d41e0,DBLP:conf/iclr/HintonSF18} have explored a class of networks called Capsule networks which use a group of neurons (``capsule'') to explicitly represent not only the presence of an object but also parameters such as position and orientation. More recently, Hinton \cite{DBLP:journals/corr/abs-2102-12627} has proposed an ``imaginary system'' called GLOM to overcome some of the limitations of capsule networks. Independently, Hawkins and colleagues \cite{10.3389/fncir.2019.00022} have taken inspiration from neuroscience, specifically cortical columns and grid cells, to propose that the brain uses object-centered reference frames to represent objects, spatial environments and even abstract concepts.

What has been missing is a scalable framework that solves the following problem: how can neural networks learn intrinsic references frames for objects and parse visual scenes into part-whole hierarchies by dynamically allocating nodes in a parse tree? Here we introduce Active Predictive Coding Networks (APCNs), a class of structured neural networks inspired by the neocortex that address this problem using hypernetworks \cite{DBLP:conf/iclr/HaDL17} to learn and dynamically generate parse trees from images.

APCNs contribute to a number of lines of research that have not been  connected before:

\begin{itemize}
    \item \textbf{Predictive Coding:} APCNs build on predictive coding models of cortical function \cite{Rao1999PredictiveCI,friston_pc,Jiang2021DynamicPC}, which emphasize the role of hierarchical prediction and prediction errors in driving learning and inference.
    
    \item \textbf{Visual Attention Networks:}  APCNs extend previous visual attention approaches such as the Recurrent Attention Model (RAM) \cite{NIPS2014_09c6c378} and Attend-Infer-Repeat (AIR) \cite{NIPS2016_52947e0a} by learning structured  strategies for sampling the visual scene. We also define appropriate baselines for these types of models to demonstrate the utility of intelligent sampling.
    
    \item \textbf{Hierarchical Reinforcement Learning:} APCNs leverage ideas in hierarchical reinforcement learning by learning abstract macro-actions (``options'' \cite{SUTTON1999181}) to hierarchically parse an image into parse trees via hypothesis testing.
\end{itemize}
By combining predictive coding, active sampling of the visual scene and hierarchical actions, our approach also suggests a neural solution to an important challenge posed by cognitive science and AI researchers \cite{lake_ullman_tenenbaum_gershman_2017}: how can neural networks in the brain and in AI learn hierarchical compositional representations that allow new objects, scenes and concepts to be quickly created, recognized and learned?

\section{Active Predictive Coding Networks}
\label{section:apcns}
% General idea: structured representation, factored state and action networks at each level generated by embedding vectors which represent higher level state and higher level macro-action/option
Suppose there is an optical sensor with limited computational capacity connected to a device with large computational capacity via a communication channel with limited bandwidth. How can the sensor intelligently sample the scene to allow the computational device to parse and understand the scene? There is a direct correspondence between this arrangement and our visual system: the sensor is the retina, the device the visual cortex and the channel the optic nerve. As opposed to typical CNNs in AI, humans rely on intelligent sequential sampling of visual scenes via eye movements (``saccades''). The Recurrent Attention Model (RAM) \cite{NIPS2014_09c6c378} and related approaches \cite{NIPS2016_52947e0a,ba2014multiple} emulate this idea by utilizing a ``glimpse sensor'' that extracts high-resolution information about small parts of a larger input image; this information is conveyed to artificial neural networks for further processing. For example, in the RAM model, a ``location network'' decides on which location in the image to sample a glimpse from, and a recurrent neural network (RNN) integrates the sampled information for downstream tasks. Since the glimpse sensor used is not differentiable, the location network is trained via the reinforcement learning algorithm REINFORCE \cite{NIPS1999_464d828b,NIPS2014_09c6c378}.

Here we introduce active predictive coding networks (APCNs), which build on ideas explored by RAM and other models in the following ways:
\begin{itemize}
    \item Information from glimpses are organized in a structured, hierarchical way using intrinsic reference frames computed by a hierarchical network.
    \item Inspired by the formalism of Partially Observable Markov Decision Processes (POMDPs) \cite{10.5555/864920,rao2010decision}, each level of the hierarchical network is composed of a state network and an action network. The state network at each level integrates the information from input samples and implements the state transition model for POMDPs at a particular level of abstraction. The action network at each level is task specific and responsible for planning actions at that particular level of abstraction.
    \item At every hierarchical level, the state network is trained via predictive coding, while the action network is trained via REINFORCE.
\end{itemize}

\subsection{Basic Idea}
At each level of a hierarchy, the APCN model uses two embedding vectors, one to represent the current ``state'' denoting an object/part, and the other to represent the current action denoting the position (or more generally, the transformation) of the object/part. Nonlinear functions (implemented as hypernetworks \cite{DBLP:conf/iclr/HaDL17}) are used to map these vectors to lower level state transition and action functions, which act as ``programs'' to parse various parts/sub-parts via sequences of sampled locations/transformations. This process can be repeated for an arbitrary number of levels.  

\begin{figure*}[t]
\centering
\includegraphics[width=0.63\linewidth]{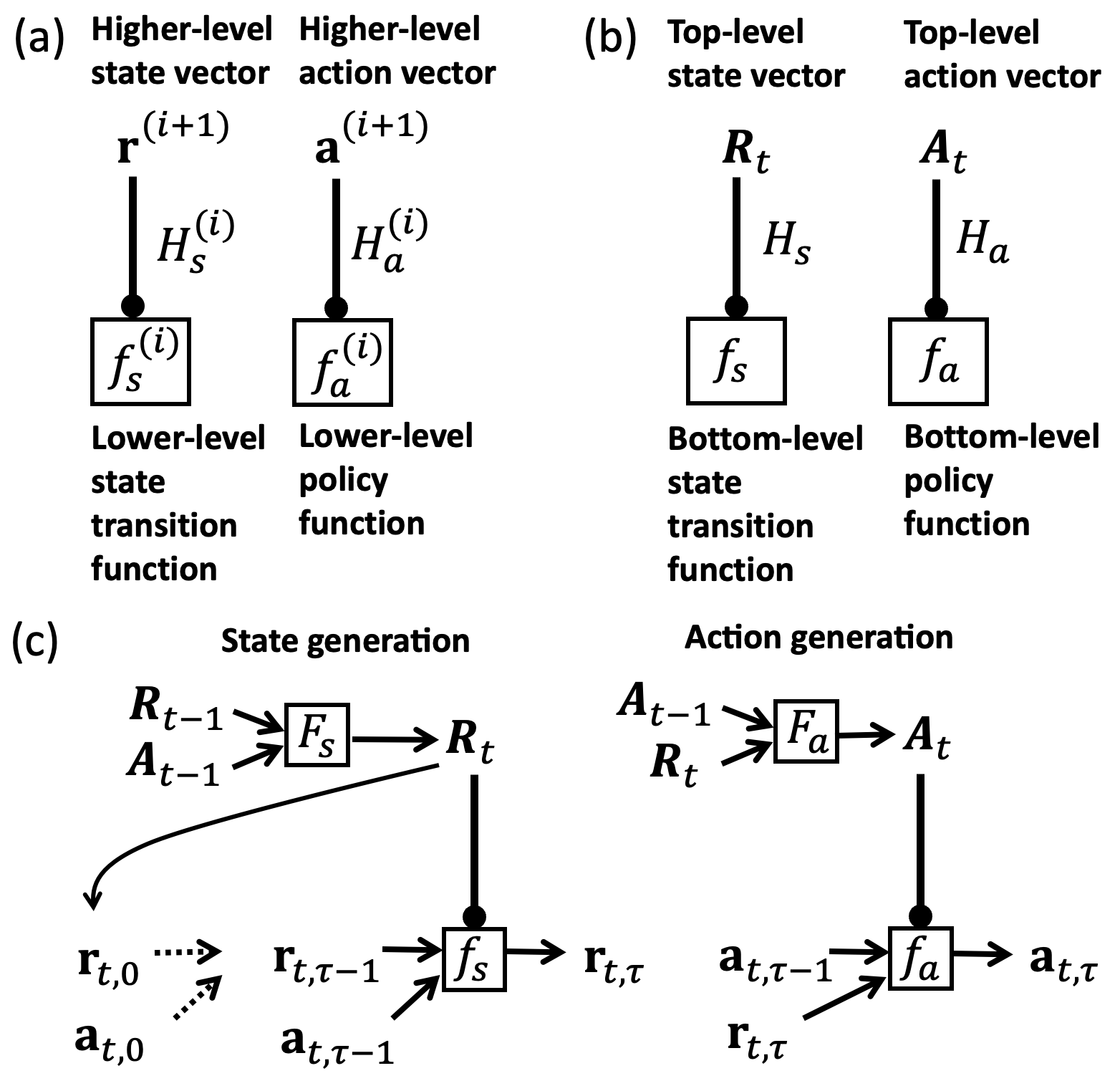}
\caption{
\textbf{Active Predictive Coding Networks (APCNs):}
    (a) Canonical generative module for hierarchical APCNs. The lower level functions are generated via hypernetworks based on the current higher level state and action embedding vectors. All functions (in boxes) are implemented as recurrent neural networks (RNNs). Arrows with circular terminations denote generation of function parameters (here, neural network weights and biases).  (b) Two-level model used in this paper. (c) Generation of states and actions in the two-level model based on past states and actions. RNNs implementing the functions in boxes additionally receive feedback from prediction errors and lower level states/actions as described in the text.
    %\textbf{(C)} Inference and learning in a two-level HCN.  State transition and policy/action functions are implemented using recurrent neural networks (RNNs). Hypernets generate lower level state and action recurrent neural networks based on current high level state and action vectors. These recurrent networks utilize prediction errors to correct their state/action estimates as they generate parts/subparts within the intrinsic reference frame defined by the higher level state vector. Prediction errors are also used to train the state transition networks. Reinforcement learning based on the task loss (here, classification loss) and the REINFORCE algorithm is used to train the action networks.
    }
\label{Fig:APCN-intro}
\end{figure*}

Figure~\ref{Fig:APCN-intro} (a) formalizes this idea and shows the canonical generative module for APCNs. The module consists of a higher level state vector ${\bf r}^{(i+1)}$, which uses a function (hypernetwork) $H^{i}_s$ to generate a lower level state transition function $f^{i}_s$ (implemented as an RNN), and a higher level action vector ${\bf a}^{(i+1)}$, which uses a function (hypernetwork) $H^{i}_a$ to generate a lower level action (or policy) function $f^{i}_a$ (also implemented as an RNN). For the present paper, we focus on a two-level model (with a top level and bottom level) as shown in Figure~\ref{Fig:APCN-intro} (b). The generation of states and actions for the two levels is shown in Figure~\ref{Fig:APCN-intro} (c). The state and action RNNs at the lower level are generated independently by their parent RNNs (via the action/state embedding vectors) but exchange information horizontally within each level as shown in Figure~\ref{Fig:APCN-intro} (c): the state network generates the next state prediction based on the current state and action while the action network generates the current action based on the current state and previous action.

%The state transition function $f^{i}_s$ takes as input the current state $r^{i}_t$ at that level at time step $t$ and the current action $a^{i}_t$ to generate the next state $r^{i}_{t+1}$ (Figure~\ref{Fig:HCN}B, "State generation"). The initial state $r^{i}_0$ is generated by the higher level state $\bf{r}^{(i+1)}$. The action network takes as input the current state $r^{i}_t$ (and optionally, the previous action $a^{i}_{t-1}$) to generate the current action $a^{i}_{t}$ (Figure~\ref{Fig:HCN}B, "Action generation"). 

In the context of parsing an image, the action vector at a given level chooses which sub-tree of the parse tree to explore next, while the state vector represents all the integrated scene information provided by the lower levels. The exploration of a sub-tree proceeds by dynamically generating state and action ``sub-programs'' for the level below via hypernetworks. In the present implementation, the lower level RNNs execute for a fixed number of time steps, generating parts and their locations, before returning control back to the higher level.\footnote{Future implementations will explore the use of termination functions \cite{SUTTON1999181,NIPS2016_52947e0a} to allow a variable number of time steps at each level and for each example.} The higher level state then transitions to the next state (the next object/part) using that level's transition function $F_s$ and the action specified by the higher level policy $F_a$, and the process continues.

\subsection{Parse Trees, Reference Frames and the Glimpse Sensor}

\begin{figure*}
\centering
\includegraphics[width=0.5\linewidth]{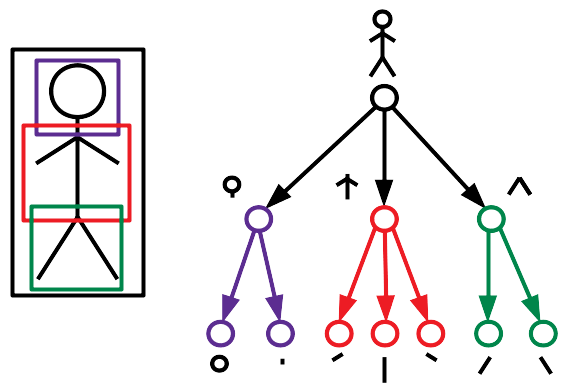}
\caption{
\textbf{Example of a Parse Tree for a Human Body:} Black box: Frame of reference for the entire object. Purple, red, green boxes: Frames of reference for the head, upper and lower body respectively. The parse tree on the right shows the part-whole hierarchy with respect to these reference frames. The leaves of the tree correspond to the  lowest-level parts of the object. Locations for all parts are computed within the parent node's reference frame, e.g., the locations for the head, upper body and lower body are computed with respect to the body's reference frame (black box).
    }
\label{fig:ontology}
\end{figure*}

Figure \ref{fig:ontology} depicts an example of a simple parse tree for a human body. We can think of the union of all these parse trees for all potential scenes as a graph representing a structured ontology of ``parent-child'' relations. This graph is hierarchical and consists of different layers, with connections between them representing part/sub-part relations. An APCN explores this ontology graph by testing different branches at each layer and extracts an appropriate parse tree for the scene.

\begin{figure*}[t]
\centering
\begin{subfigure}{0.45\textwidth}
\centering
\includegraphics[width=\linewidth]{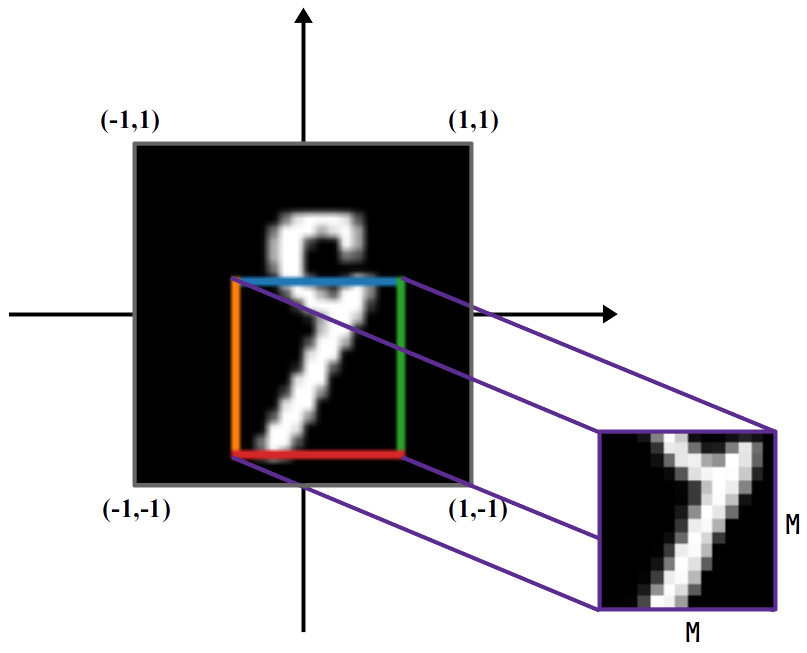}
    \caption{}
    \label{fig:sensor}
\end{subfigure}
\begin{subfigure}{0.45\textwidth}
\centering
\includegraphics[width=\linewidth]{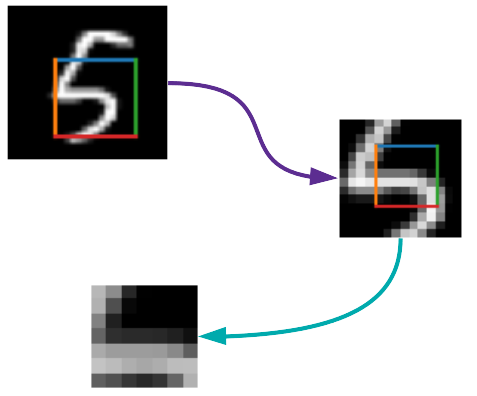}
\caption{}
    
\label{fig:frames}
\end{subfigure}\caption{
\textbf{Reference Frames for Images:} (a) The top level of the model picks a location and focuses on a sub-region of a pre-specified size (here, $M \times M$) centered at that location. This sub-region contains a higher-level part of the object at a location relative to the original reference frame of the object (here, the digit ``9''). This sub-region selected by the higher level in turn acts as the reference frame for the lower level. (b) Two-level model. The top level focuses on a sub-region $\frac{1}{4}$ the area of the initial input and fixes this region as the current reference frame. The next level focuses on  locations within this local frame of reference and extracts sub-sub-regions, which contain sub-parts of the higher level part at particular locations, all calculated relative to the local frame of reference. This hierarchical factoring of parts, sub-parts and their transformations within local reference frames is critical for compositionality.
    }
\label{fig:interaction}
\end{figure*}

Parse trees for images imply spatial convergence as one goes up the representational hierarchy since typically an entity has a larger spatial extent than its constituent parts. APCNs implement this idea using recursive object-centered reference frames. The top level of an APCN architecture spans the entire image. At each step, the network chooses a sub-region of the image to focus on (Figure \ref{fig:sensor}). It then generates a lower-level parser (comprised of state-action sub-networks) and assigns this image sub-region as the input. The bottom-most level has direct access to the image via small-sized glimpses. APCNs perform a type of depth-first exploration of the representational graph where each layer descends deeper into the graph with a new object-centered reference frame. These stacks of reference frames can be composed to derive the absolute location of any sampled glimpse within the image. Figure \ref{fig:frames} shows an example of recursive reference frame traversal down a two-level hierarchy.

 Interactions with an image $I$ (of size $N\times N$ pixels) are carried out through a glimpse sensor $G$. This sensor takes in a location $l$ and a fixed scale fraction $m$, and extracts a square glimpse/patch $g = G(I,l,m)$ centered around $l$ and of size $(mN)\times(mN)$. Since $l$ is continuous, the sensor is implemented using a differentiable bilinear interpolation module as introduced in \cite{NIPS2015_33ceb07b}. The image dimensions are normalized so that $l \in [-1,1]$ and $m$ is hard-coded for each layer. Other transformations such as rotation and shear are ignored in the current version of our model but present an obvious direction for future research. 

\subsection{Inference in the Active Predictive Coding Network}
Without loss of generality, we consider a two-level version of the APCN architecture in the current paper, with the understanding that it can be easily extended to more levels. The two levels operate at different time scales. For a given input, the top level runs for $T_2$ steps which we will refer to as ``macro-steps.'' For each macro-step, the bottom level runs for $T_1$ ``micro-steps,'' yielding a total of $T=T_2 T_1$ steps for the entire network to process each input. Let $F_s,F_a$ be the top level state and action networks respectively. Let $R_{t},A_{t}$ be the recurrent activity vectors of these networks (i.e., the top level state and action vectors) at macro-step $t$. Let $f(;\theta)$ denote a network parameterized by $\theta$. In the case of a fully connected network with $L$ layers, $\theta = \{W_l,b_l\}_{l=1}^{L}$ contains the weight matrices and biases for all the layers. The state and action RNNs of the bottom level are denoted by $f_s(;\theta_{s})$ and $f_a(;\theta_{a})$ while their activity vectors are denoted by $r_{t,\tau}$ and $a_{t,\tau}$ respectively, where $t$ ranges over macro-steps and $\tau$ over micro-steps. Figure~\ref{Fig:APCN-intro}C depicts these RNNs and activity vectors.

\begin{figure*}[t]
\centering
\begin{subfigure}{0.4\textwidth}
\centering
\includegraphics[width=\linewidth]{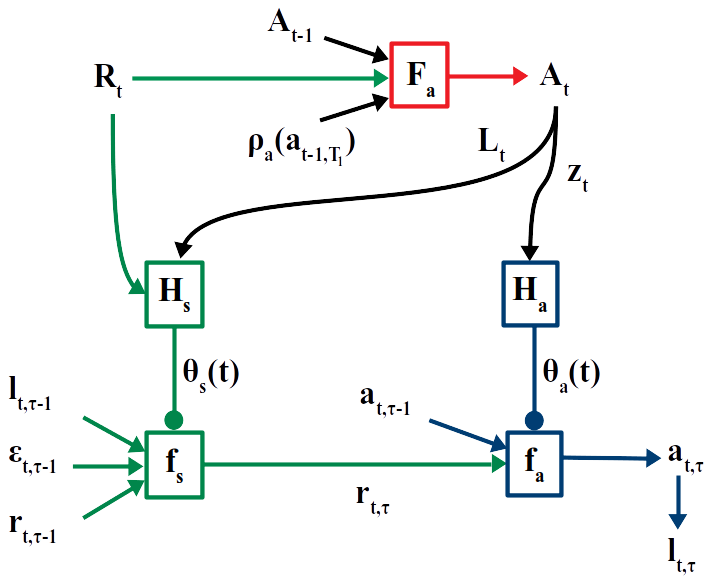}
    \caption{}
    \label{fig:main-hyper}
\end{subfigure}
\begin{subfigure}{0.3\textwidth}
\centering
\includegraphics[width=\linewidth]{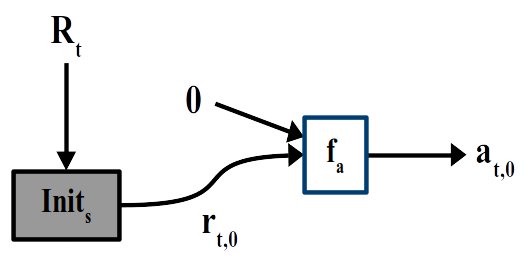}
    \caption{}
    \label{fig:main-init}
\end{subfigure}\\
\begin{subfigure}{0.3\textwidth}
\centering
\includegraphics[width=\linewidth]{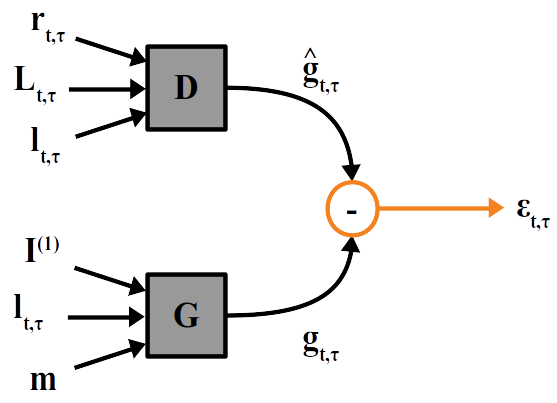}
    \caption{}
    \label{fig:main-prediction}
\end{subfigure}
\begin{subfigure}{0.4\textwidth}
\centering
\includegraphics[width=\linewidth]{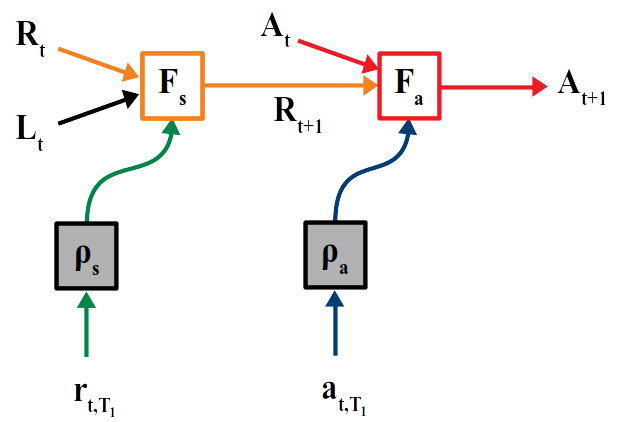}
    \caption{}
    \label{fig:main-feedback}
\end{subfigure}

    \caption{
        \textbf{Inference in Active Predictive Coding Networks:} (a) Dynamic generation of bottom-level state RNN $f_s$ and action RNN $f_a$ (``sub-programs'') from top-level state vector $R_t$ and action vector $A_t$. This diagram elaborates the one in Figure~\ref{Fig:APCN-intro} (c) for the case of parsing images, showing how actions produce locations and options, and how prediction errors and feedback from the lower level are used to update state vectors at the two levels. See text for details.  (b) Initialization of bottom-level state by the higher-level state $R_t$ via a network $\text{Init}_s$. (c) Computation of prediction error between predicted glimpse $\hat{g}$ generated by decoder $D$ for the current time step $({t,\tau})$ and actual glimpse image $g$ from the glimpse sensor after moving to the location $l_{t,\tau}$ produced by the action network. (d) Update of top-level state $R_t$ and action $A_t$ based on feedback (via networks $\rho_s$ and $\rho_a$) upon bottom-level sub-program termination (after $T_1$ micro-steps).
    }
\label{fig:main}
\end{figure*}

\subsubsection{Higher-Level Network Operation}
At each macro-step $t$, the top level action RNN updates its activity vector to $A_t$ which generates two values: (a) a location $L_t$ and (b) a macro-action (or option) $z_t$ (Figure~\ref{fig:main-hyper}). The location $L_t$ is used to restrict the bottom level to a sub-region $I^{(1)}_t = G(I,L_t,M)$ corresponding to a new frame of reference of scale $M$, centered around $L_t$ (Figure \ref{fig:sensor}). The option $z_t$ is used as an embedding vector input to a non-linear function, implemented by a hypernetwork $H_a$, to dynamically generate the parameters $\theta_{a}(t) = H_a(z_t)$ of the lower-level action RNN. For exploration during reinforcement learning, we treat the output of the location network as a mean value $\bar{L}_t$ and add Gaussian noise with fixed variance to sample an actual location: $L_t = \bar{L}_t + \epsilon$, where $\epsilon \sim \mathcal{N}(0,\sigma^2)$. We do the same for the option $z_t$.

The state vector $R_t$ and location $L_t$ are fed as inputs to the state hypernetwork $H_s$ to generate the parameters $\theta_{s}(t) = H_s(R_t,L_t)$ specifying a dynamically generated bottom-level state RNN for the current frame of reference. Figure \ref{fig:main-hyper} illustrates this top-down generation process.

\subsubsection{Lower-Level Network Operation}
At the beginning of each micro-step, the higher-level state $R_t$ is used to initialize the bottom-level state vector via a small feedforward network $\text{Init}_s$ to produce $r_{t,0} = \text{Init}_s(R_t)$ (Figure \ref{fig:main-init}). Each micro-step proceeds in a manner similar to a macro-step. The bottom-level action RNN updates its activity vector $a_{t,\tau}$ based on the current state and past action, and a location $l_{t,\tau}$ is chosen as a function of $a_{t,\tau}$ (Figure \ref{fig:main-hyper}, lower right). This results in a glimpse image $g_{t,\tau} = G(I^{(1)}_t,l_{t,\tau},m)$ of scale $m$ centered around $l_{t,\tau}$ within image sub-region $I^{(1)}_t$ specified by the higher level (Figure \ref{fig:main-prediction}). The frames of reference and the corresponding image sub-regions across the two levels are depicted in Figure \ref{fig:frames}.

To predict the next glimpse image at the location specified by the action network, the lower-level state vector $r_{t,\tau}$, along with locations $L_t$ and $l_{t,\tau}$, are fed to a generic decoder network $D$ to generate the predicted glimpse $\hat{g}_{t,\tau}$ (Figure \ref{fig:main-prediction}). This predicted glimpse is compared to the actual glimpse image to generate a prediction error $\epsilon_{t,\tau} = g_{t,\tau} - \hat{g}_{t,\tau}$. Following the predictive coding model \cite{Rao1999PredictiveCI}, the prediction error is used to update the state vector via the state network: $r_{t,\tau+1} = f_s(r_{t,\tau},\epsilon_{t,\tau},l_t;\theta^{(s)}(t))$ (Figure \ref{fig:main-hyper}, lower left). For the bottom-level locations, we follow the same Gaussian noise-based exploration strategy as the top-level.

At the end of each macro-step (after $T_1$ bottom-level micro-steps have finished executing), the top level state RNN activity vector is updated using the final bottom-level state vector and the top-level location: $R_{t+1} = F_s(R_t,\rho_s(r_{t,T_1}),L_t)$ where $\rho_s()$ is a single-layer state ``feedback'' network. Figure~\ref{fig:main-feedback} (left side) depicts this process. The top-level action RNN activity vector $A_t$ is then updated using $R_{t+1}$ and $\rho_a(a_{t,T_1})$ (Figure~\ref{fig:main-feedback} (right side)), and the process continues. The above steps correspond to a sub-program in the state/action hierarchy terminating and returning its result to be integrated by its parent. Note that this architecture can be readily extended to more levels by having $F_s,F_a$ be dynamically generated by another parent level, and so on.

\subsection{Training the Active Predictive Coding Network}
The state and action networks are trained separately via different loss functions. The state networks are trained to minimize prediction errors via backpropagation while the action networks are trained to minimize total expected task loss via REINFORCE together with backpropagation. During training, whenever the state vectors at any given level are passed as input to that level's action network (see Figure~\ref{fig:main-hyper}), the gradients for backpropagation are cut off. The goal of the state prediction network is to predict the next state and is task-agnostic. The goal of the action network is to choose effective actions given past states and actions, so that the task loss is minimized. 

\subsubsection{State Networks}
\label{model:state-system}
%
% What parameters are trained here?
%

The prediction error $\epsilon_{t,\tau}$ is given by:
\begin{equation}
    \epsilon_{t,\tau} = g_{t,\tau} - \hat{g}_{t,\tau} = G(I^{(1)}_t,l_{t,\tau},m) - D(r_{t,\tau},L_t,l_{t,\tau})
\end{equation}
The prediction error loss function is given by:
\begin{equation}
    L_\text{pred} = \sum_{t=1}^{T_2}\sum_{\tau=1}^{T_1} \| \epsilon_{t,\tau} \|^2_2
\end{equation}
At the end of a macro-step $t$, the higher level  also reconstructs the current reference image $I^{(1)}_\text{ref}$, downsampled to the size of a lower-level glimpse, using a decoder $D_\text{ref}$ with inputs $R_{t+1}$ and $L_t$, yielding the loss function $L_\text{ref} = \sum_{t=1}^{T_2}\|I^{(1)}_\text{ref} - D_\text{ref}(R_{t+1},L_t) \|_2^2$. The total loss function for training the state networks at the two levels via backpropagation is given by:
\begin{equation}
L_\text{state} = L_\text{pred} + L_\text{ref}
\end{equation}

\subsubsection{Action Networks}
To apply APCNs to a given task (such as image reconstruction or classification), either the state or action RNN vectors can be provided as input to another neural network trained for the task. Here we use the action vectors. Let $A_\text{out}(t,\tau) = [A_{t}$~$a_{t,\tau} ]^T $ be the concatenation  of top- and bottom-level action vectors for time step $(t,\tau)$. Let $L_\text{task}$ be the task loss.
%We can train the action networks via back-propagation on this loss. Since there are no gradients passed from the state networks to the action networks, the location networks at both levels do not have gradients with respect to $L_\text{task}$. We combine the task loss with the REINFORCE algorithm to provide gradients for both networks.
Using just the final $A_\text{out}$ (as in RAM \cite{NIPS2014_09c6c378}) for training actions has the shortcoming that the resulting reward function is sparse (the model is evaluated only after the final step). We use a dense, structured reward function (in our case, a dense loss function) as follows. For each micro-step, we compute the marginal change in loss after the action for that step (i.e., fixating on a new location) has been executed:
\begin{equation}
R_{t,\tau} =  L_\text{task}(A_\text{out}(t,\tau-1)) - L_\text{task}(A_\text{out}(t,\tau))
\end{equation} 
For example, if the task is reconstruction of an image, the reward is positive if the new action (new fixation location) reduced the reconstruction error.

For each macro-step, we compute the marginal change in loss due to the whole macro-step:
\begin{equation}
R_{t} =  L_\text{task}(A_\text{out}(t-1,T_1)) - L_\text{task}(A_\text{out}(t,T_1))
\end{equation} 

The top layer is trained using the cumulative reward from all future macro-steps $\Phi_t = \sum_{i=t}^{T_2}R_i$, whereas the bottom layer is trained using the future rewards within each macro-step $\Phi_{t,\tau} = \sum_{j=\tau}^{T_1} R_{t,j}$. This corresponds to the intuition that micro-actions taken inside different frames of reference should not affect each other in terms of reward.

We use an adjusted version of the baseline-based variance reduction technique introduced in \cite{NIPS1999_464d828b} and used in \cite{NIPS2014_09c6c378}. We learn two separate baselines: $b_{t,\tau} = \mathbb{E}[\Phi_{t,\tau}]$ and $b_t = \mathbb{E}[\Phi_{t}]$ and use the baseline-removed cumulative rewards  $\Phi_{t,\tau}-b_{t,\tau}$ and $\Phi_{t}-b_{t}$ for training.

The REINFORCE loss is given by:
\begin{equation}\label{eq:REINFORCE}
L_\text{REINFORCE} = J(\theta_\text{L},\theta_\text{l}) = -\sum_{t=1}^{T_2}\underbrace{\left(\log P(L_t | A_t ; \theta_L)(\Phi_t-b_t)+ \sum_{\tau=1}^{T_1}\log P(l_{t,\tau} | a_{t,\tau} ; \theta_l) (\Phi_{t,\tau}-b_{t,\tau})\right)}_\text{Action log-probabilities}
\end{equation}
% Define theta_loc
% Reinforce only for loc network?
% Explain why only location parameters being trained
% Explain how log probabilities are computed - location network details
% What parameters are being trained 

As mentioned earlier, to allow exploration during training with  REINFORCE, the locations at each macro- or micro- step were the location network's output plus Gaussian noise. Therefore the logarithmic probability terms above reduce to the squared Euclidean distances between the mean and the sampled locations.

%Motivate combined loss better

We combine the REINFORCE loss with a  dense version of the task loss to get the combined loss function for the action networks:
\begin{equation}
    L_\text{action} = \underbrace{L_\text{REINFORCE}}_\text{Location networks} + \underbrace{\sum_{t}\sum_{\tau} L_\text{task}(A_\text{out}(t,\tau))}_\text{Action sub-system minus location networks}
\end{equation}
For example, if the task is reconstruction, the second term in the combined loss allows minimization of the reconstruction error at every time step. Overall, the combined loss function increases the performance of the intermediate action vectors from step to step in the context of the task, producing more interpretable results. To encourage the action networks to produce locations within image boundaries, locations were regularized using a soft $\ell_2$ penalty (see Appendix \ref{app-penalty}).
% location penalty for both levels?

\section{Results}
Our first set of experiments compares the performance of ACPNs to baseline methods. The second set of experiments demonstrates the ability of APCNs to learn part-whole hierarchies. The third set of experiments evaluates the compositionality of learned APCN representations in a transfer learning task.

\subsection{Comparison with Baselines}
\subsubsection{Baseline 1: Random Policy}
To evaluate whether the learned policies in APCNs are ``intelligent,'' it is crucial to compare them with appropriate baselines. A simple baseline policy is randomly sampling different glimpse locations. We refer to this as the Randomized Baseline model with $T$ glimpses or RB(T). We sample $T$ i.i.d. locations $\{l_t\}_{t=1}^{T}$ from a box of height and width such that the whole glimpse $g_t$ resides within the boundaries of the image. Each glimpse and its location are concatenated and passed through a feature extractor $F$ to obtain a feature vector $f_t = F([g_t,l_t])$. The $T$ feature vectors are averaged to obtain the latent vector $\bar{f} = \frac{1}{T}\sum_{t=1}^{T} f_T$. This vector is given as input to a feedforward network that is trained for the task.

The authors in \cite{NIPS2014_09c6c378} considered a baseline that consists of the RAM model applied to a single glimpse randomly sampled from the whole image. This baseline achieved $57.15\%$ accuracy on the MNIST classification task. In our case, the RB(3) baseline achieved $93.1\%$ classification accuracy.

%$92.3\%$
Given this strong classification performance of RB(3), we conclude that simple datasets such as MNIST may be unsuitable for evaluating the performance of intelligent sampling (attention) models in classification tasks since a few random glimpses are sufficient to achieve reasonably high accuracy without any intelligent strategy. Our results also suggest that the ``intelligent sampling'' in RAM-like model for MNIST may be spurious, having  no major impact on classification performance.
%However, this is not an issue for the clutteredMNIST dataset, as seen in Table \ref{table:baseline-accs}, where random glimpses are unlikely to find the digit and intelligent sampling of the visual scene makes difference.

These results also suggest that rather than classification, the task of reconstructing an object, such as an MNIST digit, might be a more appropriate task for learning and enumerating parts of an object. We therefore use image reconstruction as the task for evaluating APCNs.  

\subsubsection{Baseline 2: Single level APCN}
Our second baseline is a single level version of our APCN model, which is similar to the original RAM model except: (a) instead of a single RNN, there is a separation into an action RNN responsible for the task and a state RNN that integrates glimpse information; (b) the state network is trained via predictive coding applied to predicting the next  glimpses as described in Section \ref{model:state-system}; (c) we use dense rewards to improve training and interpretability. Note that all of the above are novel additions to the original RAM model, enabling the new model to: (a) re-use the state network for multiple tasks, and (b) make the contribution of each glimpse interpretable. We will refer to this model as APCN-1 and to the two layer APCN as APCN-2. Results  comparing APCN-2 to the two baselines on the reconstruction task are described in the next section.
%
% Complete this section with results
%

\subsection{Learning Part-Whole Hierarchies via Active Predictive Coding}
We applied APCNs to the task of part prediction and reconstruction of objects in the following datasets: (a) MNIST: Original MNIST dataset of 10 classes of handwritten digits \cite{lecun-mnisthandwrittendigit-2010}.
    %\item Rot-MNIST: MNIST digits randomly rotated at $10$ distinct angles equally spaced between $[-\frac{\pi}{2},\frac{\pi}{2}]$.
(b) Fashion-MNIST: Instead of 10 digits as in MNIST, the dataset consists of $10$ classes of clothing items \cite{xiao2017fashionmnist}.
(c) Omniglot: $1623$ hand-written characters from $50$ alphabets, with $20$ samples per character \cite{a855d9fc35fa423ab5022fec85de0d6e}.

For all APCN models, a single dense layer, together with an initial random glimpse are used to initialize the state and action vectors of the top level. More experimental details for this section are discussed in Appendix \ref{app-exp-details}.

\begin{table}[]
\center
\begin{tabular}{|l|l|l|l|l|l|}
\hline
   & \textbf{MNIST} & \textbf{Fashion MNIST} & \textbf{Omniglot-Test}& \textbf{Omniglot-Transfer} \\\hline
\textbf{RB}    & $0.0120$ & $0.0145$ & $0.0307$  & $0.0301$     \\ \hline
\textbf{APCN-1}& $0.0114$ & ${\bf 0.0138}$ & $0.0324$  & $0.0323$     \\ \hline
\textbf{APCN-2}      & ${\bf 0.0085}$ & ${\bf 0.0138}$ & ${\bf 0.0227}$  & ${\bf 0.0226}$    \\ \hline
\end{tabular}

\caption{\textbf{Reconstruction Mean-Squared-Error  (per pixel).}}
\label{table:rec-mse}
\end{table}

\subsubsection{Task Performance}
We first applied APCNs to the task of  reconstructing MNIST and Fashion-MNIST datasets. For APCN-2, we used $3$ macro- and $3$ micro-steps. A comparison of APCN-2 performance with the baselines based on test set MSE is shown in Table \ref{table:rec-mse}. Note that APCN-2 is more constrained than APCN-1 since the locations within a macro-step have to reside within the frame of reference of $I^{(1)}$. APCN-2 receives additional information in the form of $T_2$ peripheral glimpses obtained by downsampling $I^{(1)}_t$ as discussed in Section \ref{model:state-system}. These peripheral glimpses are used during training but not during inference. The fact that both APCN models perform better than the random baseline shows that intelligent sampling strategies have an effect on reconstruction task performance.

We also applied APCN-2 to reconstructing Omniglot characters using $4$ macro-steps instead of $3$. Table \ref{table:rec-mse} shows that APCN-2 outperforms the baselines on this task. 

\subsubsection{Parse Strategies and Learned Part-Whole Hierarchies}
\begin{figure*}
\centering
\includegraphics[width=\linewidth]{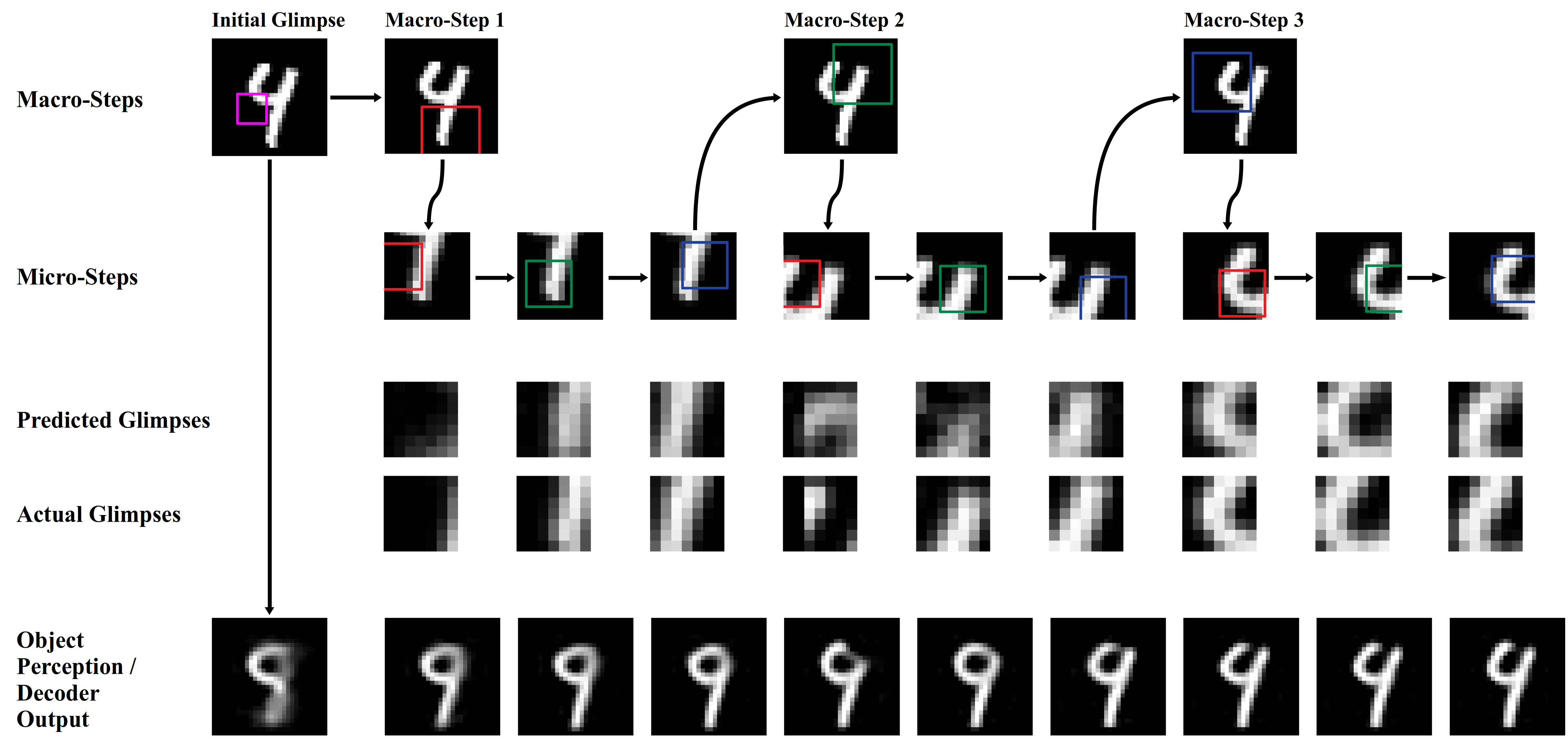}
\caption{
\textbf{Example of Learned Two-Level Parsing Strategy:} 1st row: Initialization glimpse (purple box) and sampled top-level reference frames (red, green, blue boxes), 2nd row: Sampled bottom level parts within each top-level frame, 3rd \& 4th rows: Predicted versus actual parts/glimpses, and 5th row: ``Perception'' of the model (object reconstructed from current network state) over time.
}
\label{fig:parse-example}
\end{figure*}

An example of a learned parsing strategy is shown in Figure \ref{fig:parse-example}. For each input, APCN-2 learns structured parsing strategies: the top-level learns to cover the input image sufficiently while the bottom level learns to parse sub-parts inside those sub-areas of the object.  

\begin{figure*}[t]
\centering
\begin{subfigure}{.5\textwidth}
 \centering
  \includegraphics[width=.9\linewidth]{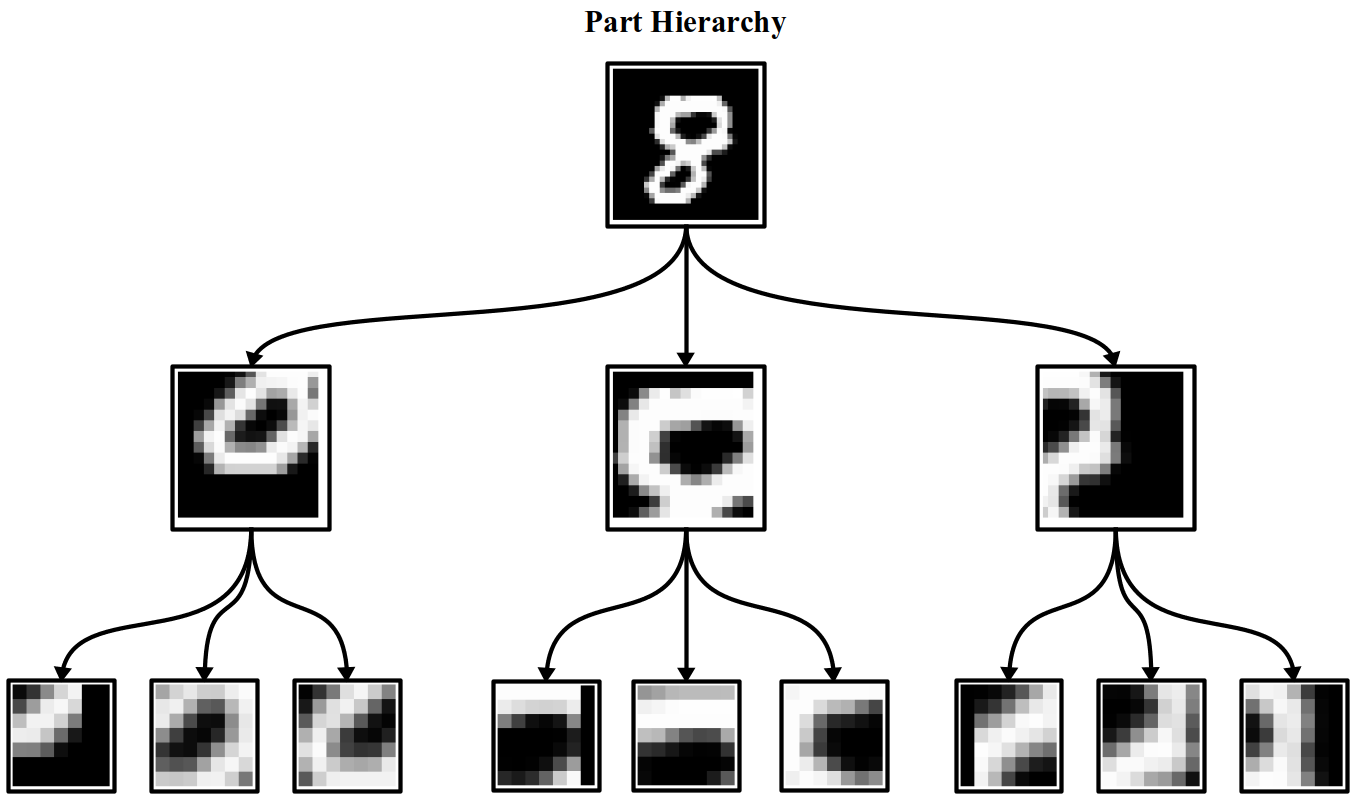}
    \caption{}
    \label{fig:parts-hierarchy}
\end{subfigure}
~
\begin{subfigure}{.5\textwidth}
 \centering
  \includegraphics[width=.9\linewidth]{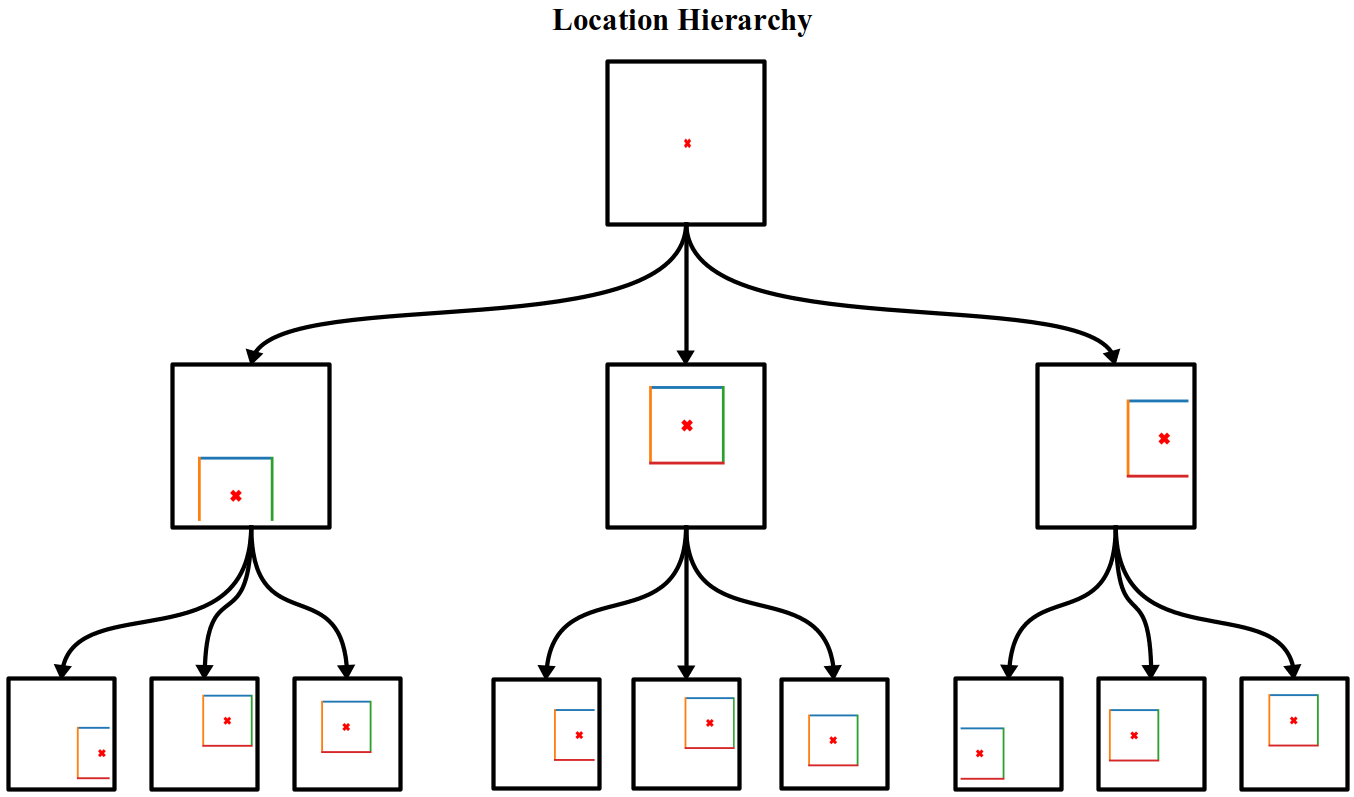}
    \caption{}
    \label{fig:locs-hierarchy}
\end{subfigure}
\caption{
\textbf{Example Parse Tree with Inferred Locations of Parts:} Hierarchy of (a) sampled parts and (b) sampled locations, inducing a hierarchy of reference frames.
}
\label{fig:parse-tree-example}
\end{figure*}

A learned part-whole hierarchy for an MNIST input, in the form of a parse tree of parts and sub-parts with locations, is shown in Figure \ref{fig:parse-tree-example}. The learned strategies sample a wide variety of parts and sub-parts of the object (strokes and mini-strokes).

\begin{figure*}
\centering
\begin{subfigure}{\textwidth}
\centering
\includegraphics[width=0.75\linewidth]{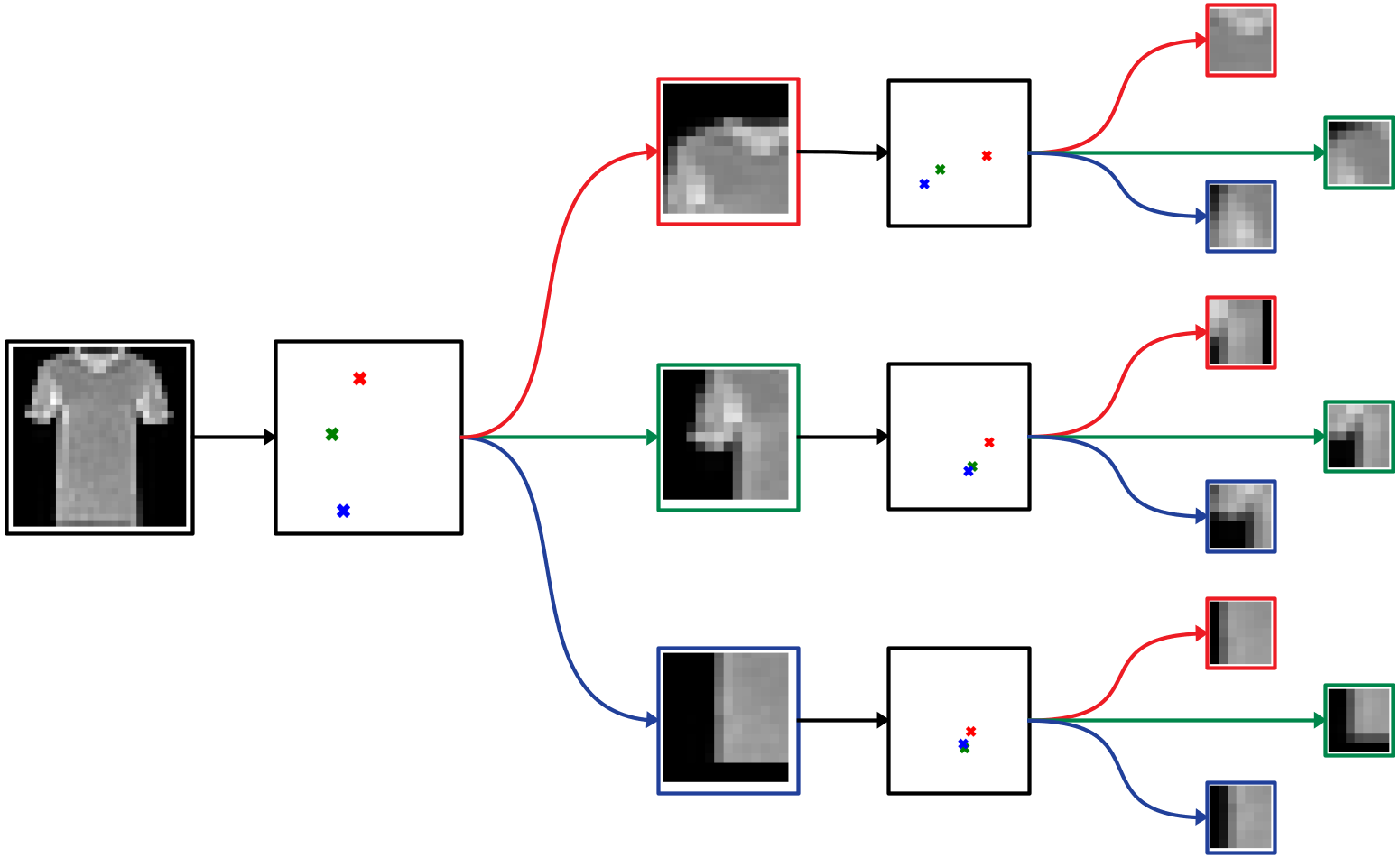}
    \caption{}
    \label{fig:class-tshirt}
\end{subfigure}\\
\begin{subfigure}{\textwidth}
\centering
\includegraphics[width=0.75\linewidth]{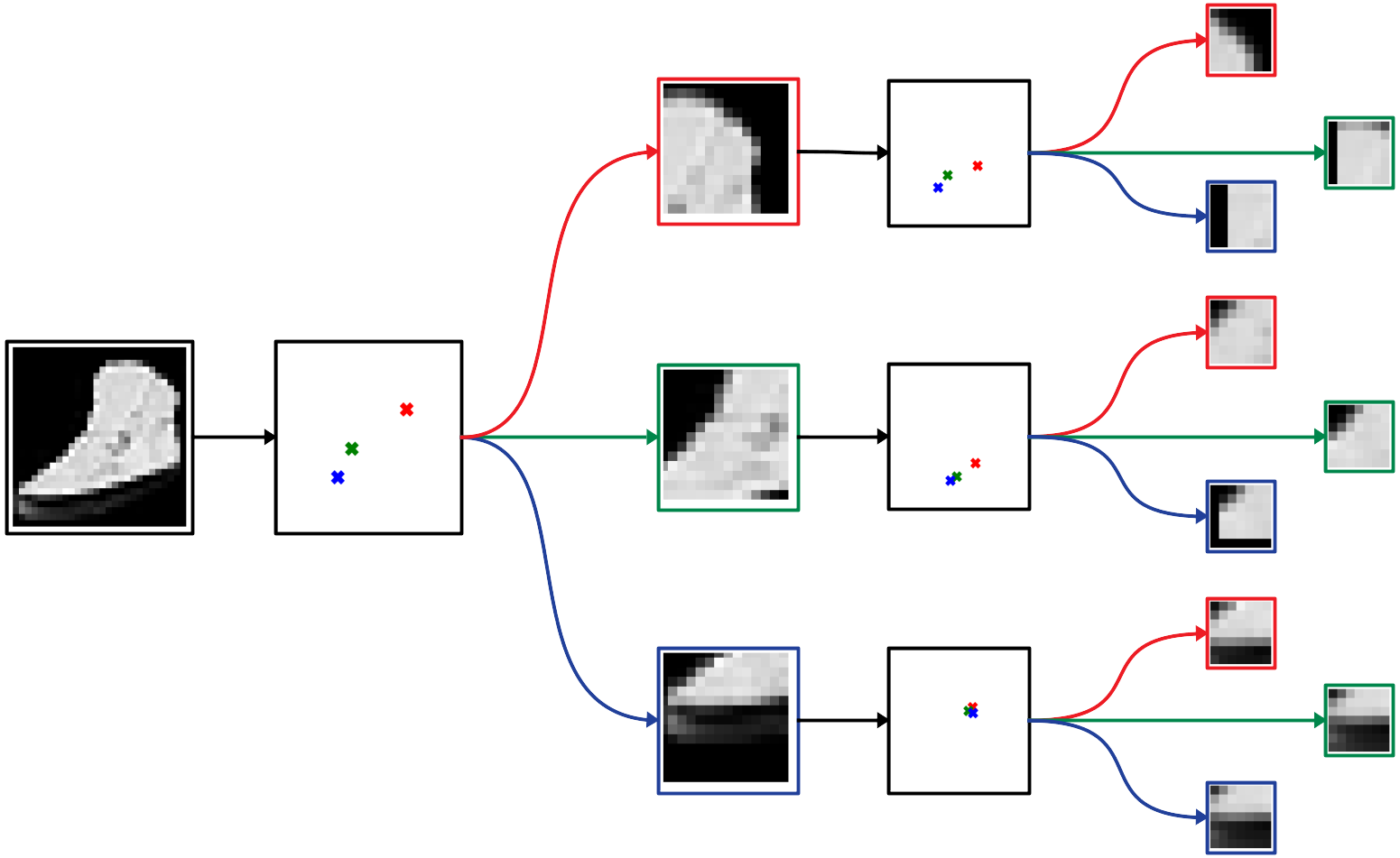}
    \caption{}
    \label{fig:class-sneaker}
\end{subfigure}\\
\centering
\caption{
\textbf{Class-Based Hierarchical Representation of Object Parts and Locations:} Average sampled locations per class, together with sampled parts for one specific example for (a) the t-shirt and (b) the sneaker classes. The order of sampled locations within each frame of reference is 1st: red, 2nd: green and 3rd: blue.
}
\label{fig:class-based-parts}
\end{figure*}

\begin{figure*}
% \begin{subfigure}{\textwidth}
% \includegraphics[width=\linewidth]{figs/mnist_clusters.png}
%     \caption{}
%     \label{fig:parts-hierarchy}
% \end{subfigure}
%\begin{subfigure}{\textwidth}
\includegraphics[width=\linewidth]{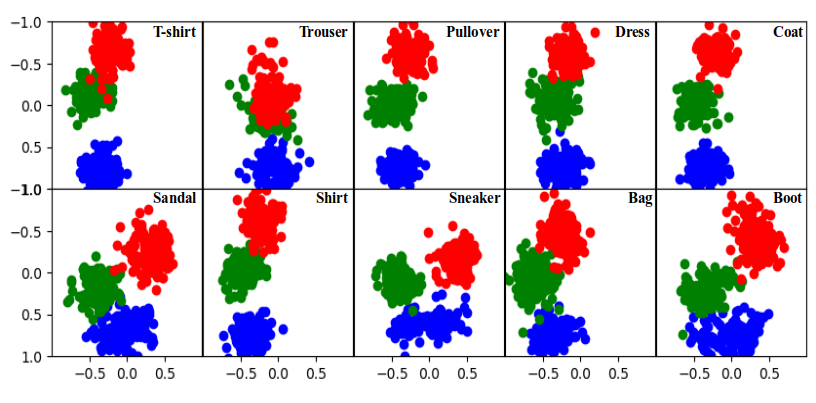}
    %\caption{}
    %\label{fig:locs-hierarchy}
%\end{subfigure}
\centering
\caption{
\textbf{Top-level Part Locations for Fashion-MNIST Examples by Class:} Red: first, Green: second and Blue: third location. Note the differences in top-level action strategies between vertically symmetric items (shirts, trousers, bags) and footwear (sandals, sneakers, boots).
}
\label{fig:top-parts-FMNIST}
\end{figure*}

An important question is whether APCN-2 learns different parsing strategies and different part-whole hierarchies for different classes of objects. Figure \ref{fig:class-based-parts} shows that this is indeed the case, for example, if we consider two different Fashion-MNIST clothing items, e.g., t-shirts versus sneakers. The figure shows that the average sampled locations of learned parts are different for these two different classes. Additional examples of learned higher-level part locations for each class in the Fashion-MNIST dataset are shown in Figure~\ref{fig:top-parts-FMNIST}.

\subsubsection{Prediction of Parts and Pattern Completion}
\begin{figure*}
\centering
\includegraphics[width=\linewidth]{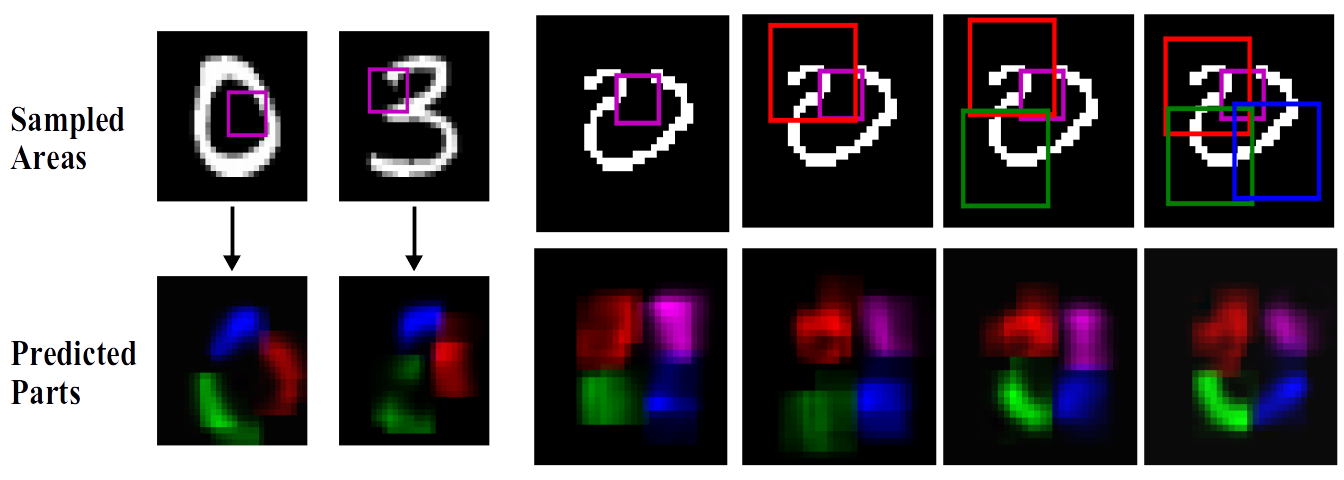}
\caption{
\textbf{Prediction of Parts and Pattern Completion by APCNs:} (Left two columns) Given only an initialization glimpse (purple box) for an input image (here, a 0 and a 3 from MNIST), the trained APCN predicts its best guess of the parts of the object and their locations (colored segments in row below). (Right columns) In other cases (here, an Omniglot character), the initial glimpse allows only a coarse prediction of parts. This prediction can be refined (bottom row) with additional glimpses (red, green, blue boxes). 
}
%\textbf{Prediction and Pattern Completion by APCNs:} Given an incomplete input, APCN predicts rest of the object by hallucinating object locations with large prediction errors and replacing those zeros (similar to robust predictive coding). MNIST, FMNIST, Omniglot examples.

\label{fig:part-prediction}
\end{figure*}

To investigate the predictive and generative ability of the model, we had the model ``hallucinate'' different parts of an object by setting the prediction error input to the lower level network to zero for all or some  macro-steps. This disconnects the model from the input, forcing it to predict the next sequence of parts and ``complete'' the object. Figure \ref{fig:part-prediction} shows that the model has learned to generate plausible predictions of parts given the initial glimpse (and any additional glimpses).

\begin{figure*}
\centering
\includegraphics[width=0.75\linewidth]{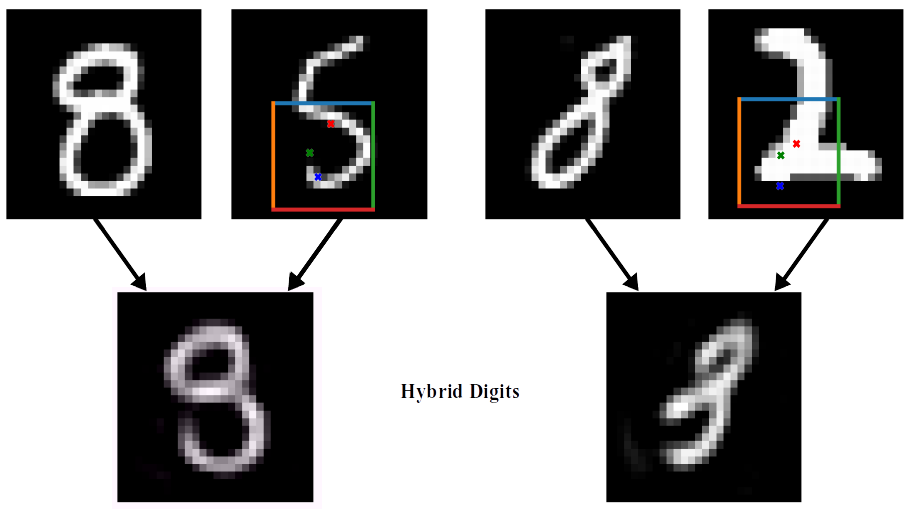}
\caption{
\textbf{Compositionality of the Learned Representations and Transfer:} Swapping the higher level state and action vectors for a pair of inputs (top row, two example input pairs: 8 and 5, 8 and 2) after two macro-steps and predicting the remaining sequence of parts results in novel objects being generated (bottom row). These results illustrate how learned ``sub-programs'' can be composed in novel ways by an APCN to facilitate transfer learning. 
}
\label{fig:compositionality}
\end{figure*}

\subsection{Compositionality and Transfer Learning}
\begin{figure*}
% \centering
% \begin{subfigure}{\textwidth}
% \centering
% \includegraphics[width=0.5\linewidth]{figs/parse_example.png}
%     \caption{Pen strokes.}
%     \label{fig:top-strategy-mnist}
% \end{subfigure}

% \begin{subfigure}{\textwidth}
% \centering
% \includegraphics[width=0.5\linewidth]{figs/super_strokes_omni.png}
%     \caption{Mini strokes.}
%     \label{fig:top-strategy-fmnist}
% \end{subfigure}
\begin{subfigure}{\textwidth}
\centering
\includegraphics[width=\linewidth]{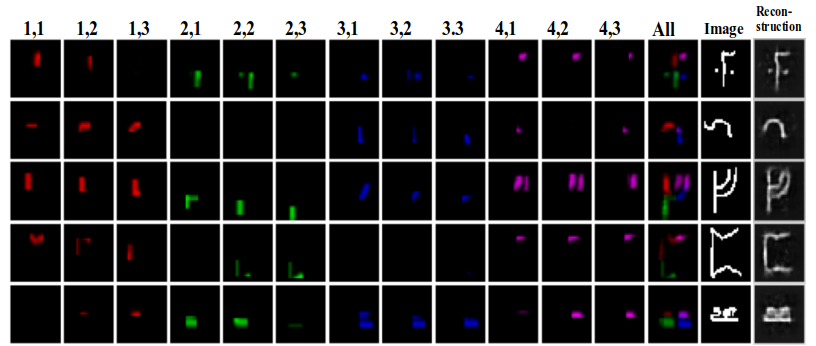}
%    \caption{Parsing novel Omniglot characters:} 
    \label{fig:top-strategy-fmnist}
\end{subfigure}

\caption{
\textbf{Transfer Learning on Unseen Classes in Omniglot:} APCNs can transfer their learned knowledge to new, previously unseen classes of Omniglot objects. Each column corresponds to parts from bottom-level glimpses at time $(t,\tau)$. The ``All'' column shows all the parts together. The last two columns show the input from the novel class and its reconstruction by the APCN.
    }
\label{fig:omniglot-parse}
\end{figure*}

Figure~\ref{fig:compositionality} illustrates the compositionality of the learned representations in an APCN. Such compositionality can be useful for transferring knowledge, in form of ``programs'' or options, from one task to another. We tested transfer learning for reconstruction of unseen character classes for the Omniglot dataset. We trained an APCN model to reconstruct examples from a subset of classes from the Omniglot alphabets. For each alphabet, we used $85\%$ of the classes for training. The rest of the classes were used to test transfer. Specifically, the trained model had to use its learned representations and programs (as generated by the hypernets) to compose and reconstruct new character classes for each alphabet. Table \ref{table:rec-mse} shows the performance of APCNs on the  transfer task. Figure \ref{fig:omniglot-parse} shows example hierarchical parsing strategies for characters from previously unseen classes, along with the reconstructions of these novel characters by the  APCN.

%\subsection{Options}

%\subsection{Ablation Study}

\section{Some Limitations of the Model}
For simple datasets such as MNIST, APCN-2 tends to converge to a general strategy that works well for all digits, resulting in little inter-class location diversity, while for other datasets such as Omniglot, a general strategy that ``tests'' a diverse set of image sub-areas might still be appropriate. For  Fashion-MNIST, different strategies are learned for vertically symmetric clothing items versus non-symmetric ones such as footwear (Figure~\ref{fig:top-parts-FMNIST}). These results can be attributed to our use of a general decoder to reconstruct the entire image based on current state within an attention-based encoder. An interesting future direction is to use a decoder where each macro-/micro- step renders a part of the object within a larger canvas used for computing the reconstruction error, as in \cite{NIPS2016_52947e0a}. 

Another limitation is the use of fixed time horizons for each option: our model parses each sub-area for a fixed number of steps whereas different sub-areas of an image might require fewer or more steps to process. Techniques such as the one used in \cite{NIPS2016_52947e0a} could be employed to address this limitation.

APCNs have yet to be applied to more challenging datasets (e.g., ImageNet) and other tasks (e.g., regression of object properties). Deeper versions of our model (with more than two levels) may be necessary for more complex image datasets.

Finally we used REINFORCE, which is inefficient compared to state-of-the-art reinforcement learning algorithms. The results could potentially be improved using more sophisticated policy gradient methods or designing novel methods tailored to the structure of the model.

\section{Conclusion}
We have presented, to our knowledge, the first hierarchical neural network capable of end-to-end learning and parsing of part-whole hierarchies from images. The framework we have proposed is highly flexible and offers a number of potential applications and future research directions. For example, actions in APCNs could include not just position but arbitrary transformations of parts, allowing the network to learn hierarchical equivariant representations, a long-sought goal in machine vision and AI \cite{DBLP:journals/corr/abs-2102-12627,DBLP:conf/iclr/HintonSF18}. More broadly, our framework offers a new approach to hierarchical reinforcement learning and planning in continuous state and action spaces. Finally, given the close connection between APCNs and predictive coding models of brain function, the proposed framework paves the way for a new interpretation of the hierarchical architecture of the cortex and a new role for cortical feedback connections in modulating the dynamics of lower-level networks \cite{Jiang2021DynamicPC} similar to the role played by hypernets in APCNs.  
%
% Generalizes to other modalities such as touch and 3D locations
% 3D reference frames or even N-D
% 
% Extends naturally to videos and action-conditioned sensory sequences

% First glimpse enough for identifying object similar to human vision

% Cite our own related work such as what-where, Lie groups

\section{Acknowledgments}
This material is based upon work supported by the Defense Advanced Research Projects Agency (DARPA) under Contract No. HR001120C0021, 
National Science Foundation (NSF) Grant \#EEC-1028725, a Weill Neurohub Investigator grant and a grant from the Templeton World Charity Foundation. The opinions expressed in this publication are those of the authors and do not necessarily
reflect the views of the funders.

\bibliographystyle{plain}
\bibliography{apcn.bib}

\appendix
\section{Implementation Details}
\label{app:details}

\subsection{Location Penalty}
\label{app-penalty}
\begin{figure*}[h]
\centering
\includegraphics[width=0.75\linewidth]{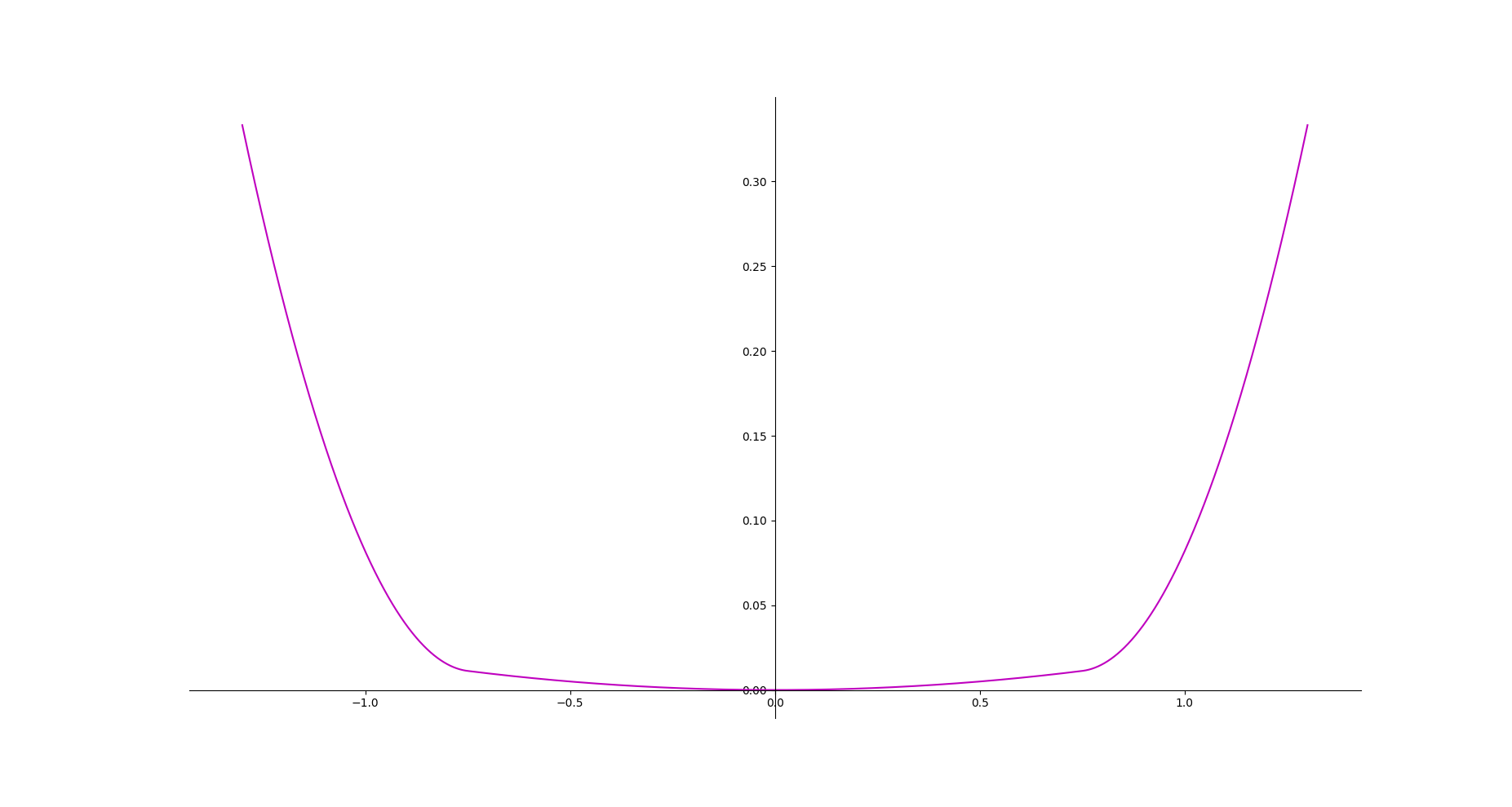}

\caption{\textbf{Location penalty function for $c=0.75$.}}
\label{fig:app-penalty}
\end{figure*}

We want to make our network avoid generating  locations exceeding the boundaries of the image. Several implementations of RAM use clipping or the hyperbolic tangent activation function. In practice, we found that constraining the locations via an appropriate penalty was more effective. We calculate a threshold $c$ so that if a glimpse is centered $c$ units away from the boundary ($l\in[-1.0+c,1.0-c]$), then the glimpse resides entirely within that boundary. We derive a thresholded version of $\ell_2$ normalization:
$$
L_\text{reg}(l) = (L_\text{Relu}(l-c))^2 + (L_\text{Relu}(-l-c))^2 - 2(\alpha c)^2
$$
The structure of this penalty can be seen in Figure \ref{fig:app-penalty}.

\subsection{Parameter Settings and Initialization}
\label{app-exp-details}
For all datasets (MNIST, Fashion-MNIST and Omniglot) the top-level action and state RNN activity vectors were of size $256$, while the lower level ones were of size $32$ for MNIST, Fashion-MNIST and $64$ for Omniglot. Both hypernetworks $H_s$ and $H_a$ consisted of four layers with sizes $256,256,64$ and $|\theta_s|$ or $|\theta_a|$. The last two layers had linear activation functions. Since the number of units of the last layer was equal to the number of parameters of $f_a$ or $f_s$, the middle layer functions as a bottleneck layer. $F_a$ and $F_s$ were implemented as RNNs with structure similar to the network used in RAM \cite{NIPS2014_09c6c378}. The option vectors $z$ were of size $32$. RELU activations were employed throughout the model apart from the last level of the reconstruction network (to avoid ``dead'' pixels). The glimpse scales were set such that $I^{(1)}$ was $14\times14$ and $g_{t,\tau}$ $7\times7$ pixels. The sizes of MNIST and Fashion-MNIST images are $28\times 28$ pixels, whereas we downsampled Omniglot characters to $32\times 32$ pixels.

For Omniglot, we reserve $85\%$ (rounded down) of the character classes in each alphabet as part of our training set. The rest of the character classes are used as the transfer set. Within the training classes, we reserve $3$ examples from each character class as part of the test set.

We utilize random glimpses to initialize the top-level state and action vectors. A random glimpse is generated at location $l_\text{init} \sim \mathcal{U}[-0.5,0.5]$. This initialization glimpse $g_\text{init}$, together with a small trainable initialization network, initializes the state vector $R_{t_{0,0}}$. The action vector is initialized using $F_a$ and $R_{t_{0,0}}$ by setting the previous action vector and feedback $\rho_A$ to all-zeroes vectors.

% \section{Additional Results}
% \subsection{Top-level locations by class}
% \label{app:results}
% \begin{figure*}
% % \begin{subfigure}{\textwidth}
% % \includegraphics[width=\linewidth]{figs/mnist_clusters.png}
% %     \caption{}
% %     \label{fig:parts-hierarchy}
% % \end{subfigure}
% %\begin{subfigure}{\textwidth}
% \includegraphics[width=\linewidth]{figs/fashion_clusters.png}
%     %\caption{}
%     \label{fig:locs-hierarchy}
% %\end{subfigure}
% \centering
% \caption{
% \textbf{Top-level part locations for Fashion-MNIST examples by class:} Red: first, Green: second and Blue: third location. Note the differences in top-level action strategies between vertically symmetric items (shirts, trousers, bags) and footwear (sandals, sneakers, boots).
% }
% \label{fig:top-parts-FMNIST}
% \end{figure*}
\end{document}